\def\year{2020}\relax
\documentclass[letterpaper]{article} % DO NOT CHANGE THIS
\usepackage{amsmath}% http://ctan.org/pkg/amsmath

\usepackage{style}  % DO NOT CHANGE THIS
\usepackage{times}  % DO NOT CHANGE THIS
\usepackage{helvet} % DO NOT CHANGE THIS
\usepackage{courier}  % DO NOT CHANGE THIS
\usepackage[hyphens]{url}  % DO NOT CHANGE THIS
\usepackage{graphicx} % DO NOT CHANGE THIS
\usepackage{graphicx}  % DO NOT CHANGE THIS
\usepackage[linesnumbered,ruled,vlined,noend]{algorithm2e}

\SetAlFnt{\small} % remove!!!!!!!!!

\SetKwInput{KwInput}{Input}                % Set the Input
\SetKwInput{KwOutput}{Output}              % set the Output
\usepackage{array}
\usepackage{xcolor}
\usepackage{xspace}
\usepackage{comment}

\newcommand{\gilad}[1]{{\textcolor{blue}{[Gilad: #1]}}}
\newcommand{\yuval}[1]{{\textcolor{red}{[Yuval: #1]}}}
\newcommand{\lior}[1]{{\textcolor{purple}{[Lior: #1]}}}
\newcommand{\MethodName}{DeepLine\xspace}

\SetCommentSty{mycommfont}

\nocopyright
\title{DeepLine: AutoML Tool for Pipelines Generation using Deep Reinforcement Learning and Hierarchical Actions Filtering}

\author{Yuval Heffetz$^1$, Roman Vainstein$^1$, Gilad Katz$^1$, Lior Rokach$^1$\\  
$^1$Software and Information Engineering Department, Ben Gurion University of the Negev\\ 
Be'er Sheva, Israel\\
\{yuvalhef, romanva\}@post.bgu.ac.il, \{giladkz; liorrk\}@bgu.ac.il 
}

\begin{document}

\maketitle

% \begin{abstract}
% Automatic machine learning (AutoML) is an area of research aimed at automating machine learning (ML) activities that currently require human experts. One of the most challenging tasks in this field is automatic ML pipeline generation: combining multiple types of ML algorithms to successfully analyze previously-unseen data. The two aspects of this task that make it difficult are its large search space and the fact that evaluating multiple pipelines is computationally expensive. In this study we present \MethodName, a reinforcement learning-based approach for auotmatic pipeline generation. Our proposed approach utilizes an efficient representation of the search space together with the leveraging of knowledge gained from previously-analyzed datasets to make the problem more tractable. Additionally, we propose a novel hierarchical action model that significantly speeds up the training of our approach. Our evaluation on 56 datasets shows that \MethodName outperforms state-of-the-art approaches both in accuracy and in computational cost.

\begin{abstract}
Automatic machine learning (AutoML) is an area of research aimed at automating machine learning (ML) activities that currently require human experts. One of the most challenging tasks in this field is the automatic generation of end-to-end ML pipelines: combining multiple types of ML algorithms into a single architecture used for end-to-end analysis of previously-unseen data. This task has two challenging aspects: the first is the need to explore a large search space of algorithms and pipeline architectures. The second challenge is the computational cost of training and evaluating multiple pipelines. In this study we present \MethodName, a reinforcement learning based approach for automatic pipeline generation. Our proposed approach utilizes an efficient representation of the search space and leverages past knowledge gained from previously-analyzed datasets to make the problem more tractable. Additionally, we propose a novel hierarchical-actions algorithm that serves as a plugin, mediating the environment-agent interaction in deep reinforcement learning problems. The plugin significantly speeds up the training process of our model. Evaluation on 56 datasets shows that \MethodName outperforms state-of-the-art approaches both in accuracy and in computational cost.

\end{abstract}

\section{Introduction}

The explosion of digital data has made the use of machine learning (ML) more ubiquitous than ever before. Machine learning is now applied to almost any aspect of organizational work, and used to generate significant value. The growth in the use of ML, however, was not matched by a growth in the number of people who can effectively apply it, namely data scientists. This shortage in skilled practitioners has spurred efforts to automate various aspects of the data scientist's work.

Automatic machine learning (AutoML) is a general term used to describe algorithms and frameworks that deal with the automatic selection and optimization of ML algorithms and their hyperparameters. Examples of AutoML include automatic hyperparameter selection for predefined algorithms \cite{hutter2011sequential}, automatic feature engineering \cite{katz2016explorekit}, and neural architecture search \cite{bello2017neural}. While effective, the above mentioned studies sought to optimize only specific steps of the overall process undertaken by human data scientists. In recent years, studies exploring the problem of \textit{automatic ML pipeline generation} have sought to automate the process end-to-end by generating entire ML pipelines.

The creation of entire ML pipelines is challenging because it involves a large and complex search space. Even simple pipelines usually involve multiple steps such as data preprocessing, feature selection, and the use of a classifier. Complex pipelines can both contain additional types of algorithms (e.g., feature engineering) and multiple algorithms from each type. The large number of available algorithms and the fact that the performance of each component of the pipeline is highly dependent on the input it receives from previous component(s) further complicates this task. 

Existing approaches for automatic pipeline generation can be roughly divided into two groups: \textit{constrained space} and \textit{unconstrained space}. Constrained space approaches generally create a predefined pipeline structure and then search for the best algorithms combination to populate it. Studies that utilize this approach include Auto-Sklearn \cite{feurer2015efficient} and Auto-Weka \cite{thornton2013auto}. This approach narrows the search space, but it prevents the discovery of novel pipeline architectures. The unconstrained space approaches place little or no restrictions on the structure of the pipeline, but they come at a higher computational cost. Approaches of this kind include TPOT \cite{olson2016tpot} and AlphaD3M \cite{drori2019automatic}.

In this study we propose \MethodName, a novel \textit{semi-constrained} approach for AutoML pipeline generation. While our approach constrains the maximal size of the pipelines, it supports the inclusion of multiple algorithm of the same type (e.g, classification), as well as the creation of parallel sub-pipelines. In addition, any compatible algorithm(s) can serve as the input of another, ensuring that novel and interconnected architectures can be discovered.

Another important advantage of \MethodName over previous work is its ability to learn across multiple datasets. We apply deep reinforcement learning (DRL) techniques that enable our approach to perform all of its learning offline. This fact considerably speeds up performance for new datasets while enabling us to leverage past experience and improve \MethodName's performance over time.

Our contributions in this study are as follows:
% \begin{itemize}
    (1) We present \MethodName, a novel approach for automatic ML pipeline generation. Our approach uses DRL to learn across multiple datasets, enabling it to efficiently produce pipelines for previously unseen datasets; (2) We propose a novel hierarchical action-modeling approach, which enables us to use fixed-size representation to model dynamic action spaces. This hierarchical solution not only speeds up the training process of the DRL agent but also enables the use of DRL methods that do not support dynamic action spaces. We implement our solution on the OpenAI Gym platform and publish the code, and; (3) We conduct an extensive evaluation on 56 datasets and show that \MethodName outperforms state-of-the-art methods, both constrained and unconstrained.

\section{Related Work}
% While our paper is dealing with the entire end-to-end ML pipeline, many of the previous works have focused on only parts of the pipeline, including the automatic optimization of hyper-parameters, for example in SMACK \cite{hutter2011sequential} or the automatic generation and selection of features, as in ExploreKit \cite{katz2016explorekit}. Some have only focused on the model selection (i.e. prediction model) part, for example using meta-learning, as done in \cite{brazdil2008metalearning}.
Applying AutoML for end-to-end pipeline generation has been an active field of research in recent years. Various studies offer a large variety of approaches to address this challenge, including Bayesian optimization \cite{hutter2011sequential} and genetic programming \cite{olson2016tpot}. One popular example is Auto-Weka \cite{thornton2013auto}, which automatically selects an algorithm for each step of a pipeline with a fixed structure and then uses Bayesian optimization (Sequential model-based optimization) to search for optimal hyperparameter settings of the pipeline. 

Following Auto-Weka, \cite{feurer2015efficient} proposed an autoML system called Auto-Sklearn. Auto-Sklearn searches through a set of pre-generated, fixed-structure pipelines. These pipelines contain \textit{placeholders} for data preprocessing, feature selection, and prediction model algorithms. It also uses past knowledge and meta-learning to guide the \textit{initial} stages of the exploration process. Auto-Sklearn also has the option to use an ensemble of its generated pipelines.

Extending the standard definition of a pipeline, TPOT \cite{olson2016tpot} uses a tree-based pipeline optimization tool for autoML. It enables the formation of dynamic pipeline architectures with multiple prediction and preprocessing algorithms that can be linked either in sequences or in parallel. TPOT uses evolutionary algorithms both for the creation of the pipeline structure and for ML algorithm selection. Hyperparameter optimization is also supported. Similarly to TPOT, Autostacker \cite{chen2018autostacker} also generates pipelines with evolutionary algorithms, but it does so using a layers-based architecture.

Recently, a different approach for pipeline generation was proposed by \cite{drori2018alphad3m}. The system, called AlphaD3M, uses a deep reinforcement learning (DRL) approach inspired by AlphaZero \cite{silver2017mastering} and expert iteration \cite{anthony2017thinking}. AlphaD3m represents the pipeline search and population challenges as a single-player game, where the player iteratively builds a pipeline by selecting from a set of actions such insertion, deletion and replacement of various pipeline elements. While these approaches enable the generation of more complex pipelines, they are also much more computationally expensive \cite{milutinovic2017end}.

While effective, all the above methods perform all their learning `from scratch' for each given dataset. Several approaches do use meta-learning, but only at a limited capacity and for initialization purposes. Our approach, on the other hand, relies heavily on learning from previously-analyzed dataset and is therefore capable to produce high-quality pipelines for new datasets at a fraction of the time previous studies require.

% \yuval{Maybe ignore alphaD3M? bc not sure if the following description is accurate since they don't provide many details in their paper}

%In this approach, the pipeline architecture is unlimited and can theoretically take any form, thus making the resulting pipelines more suitable to the problem . This is the main advantage of the latter approach over the former, but it comes on the expense of a huge search space of different architectures. We devise methods for reducing the search space to sub-spaces in each step of the model, thus making the process more feasible. 

\section{Problem Formulation}
We define a \textit{learning job} $\mathcal{L}(D, T, M)$ consisting of a tabular dataset $D$ of $m$ columns and $k$ instances, a prediction task $T$ and an evaluation metric $M$. Additionally, we define \textit{primitives} as any type of machine learning algorithm (e.g., preprocessing, feature selection, classification) and denote the set of primitives as  ${ \mathcal{P}_r = \{p_1, p_2, . . . , p_{N_p}\}}$. We define a directed acyclic graph (DAG) of primitives ${G = \{V,E\}}$, where $V \subseteq \mathcal{P}_r$ are the vertices of the graph, and $E$ are the edges of the graph and determine the primitives' order of activation. We denote $G$ as a ML pipeline, or $pipeline$ in short. 
% Given that $G$ is able to produce predictions over the specified learning job, 
Our goal is to generate a pipeline $G$ as to achieve
\[\arg\min_{G} \mathcal{E}(\mathcal{L}(D, T, M), G)\]
where $\mathcal{E}$ is the error of pipeline $G$ over the learning job $\mathcal{L}$.

To reduce the size of our search space, we consider the problem of generating $G$ as a sequential decision making task, and generate the pipeline step-by-step. We further formalize the problem as a Markov Decision Process (MDP) defined by the tuple $\mathcal{M}(\mathrm{S, A, P, R,} \gamma)$, where \textbf{$\mathrm{S}$} is the set of all possible states (i.e., pipeline and learning job configurations), $\mathrm{A}$ is the set of all possible actions defining the transitions between states, $\mathrm{P}$ is the (deterministic) transition function between states, and $\mathrm{R}$ is the reward function which is directly derived from the pipeline score with metric $M$. We also use the rewards discount factor $\gamma \in (0,1]$.

Given the expected sum of discounted rewards \( R_t=E_\pi[R(S_0)+\gamma R(S_1) + \gamma^2 R(S_2) + ...]\), produced by a policy \(\pi: \mathrm{S} \rightarrow \mathrm{A}\), our goal is to obtain
\(\arg\max_{\pi} R_t\).

Next we explain how we solve this problem through the application of reinforcement learning.
%\newline We now use this formalization to construct a reinforcement learning environment and an agent for the sequential decision making problem, assuming the Markov property is satisfied. In the following section we detail more specifically the exact methods for representing the MDP.

%Unlike some of the previous work on autoML that only focused on the automation of specific steps of the ML model, such as automatic feature generation and feature extraction, we define the problem as an end-to-end process integrating the entire model's steps, same as in TPOT \cite{olson2016tpot}, alphaD3M \cite{drori2018alphad3m}, auto-sklearn \cite{feurer2015efficient} and auto-weka \cite{thornton2013auto}. To adhere the same notations and definitions from previous works, especially \cite{olson2016tpot} and \cite{drori2018alphad3m}, we will formulate the general problem in a similar manner and define the following notations:

\section{The Proposed Framework}
Our framework consists of three main components: an \textit{environment}, an \textit{agent}, and a \textit{hierarchical-step plugin} serving as a mediator between the two. 

\subsection{The Environment}

\begin{figure*}[t]
	\centering
	\includegraphics[width=0.87\textwidth]{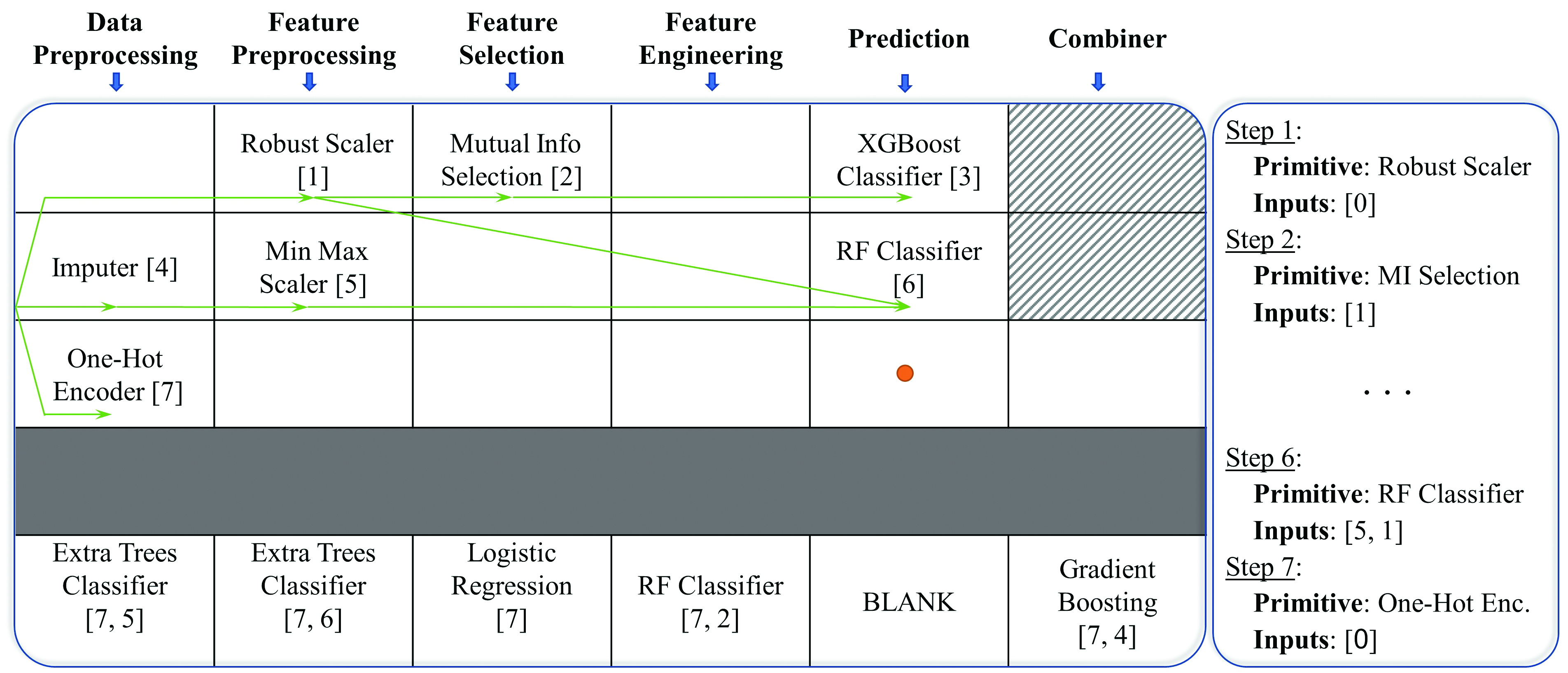}
	\caption[Grid Pipeline]{\textbf{(Left)} Example of the grid-world environment with $N$=3. The grid currently populates a pipeline with 7 steps which has 8 edges and 8 vertices (one for the raw dataset). Each column is assigned to a different primitives family. The empty cells are cells that the agent chose to populate with "blank" steps. The dot cursor automatically transitions from cell to cell, passing through all the cells of a row before moving to the next one. 
	\textbf{(Left-Bottom)} The bottom row visualize the possible actions of the current cluster, with $n$=6. Each cell in this row represents a different candidate primitive and inputs combination that the agent can choose to populate in the cell under the dot cursor.
	 \textbf{(Right)} The translation of the grid into a pipeline object, represented as a sequence of pipeline-steps.}  
	\label{fig:GridPipeline}
\end{figure*}

The main challenge we faced in defining the environment was the need to restrict the size of our state space while maintaining the ability to produce effective and complex pipelines. Our proposed solution consists of two parts: (1) the partition of the ML primitives into \textit{families}, and; (2)  a \textit{grid-world} representation of the pipeline.
% that restricts both the state space (i.e., the pipeline size) and action space.

\subsubsection{Primitive Families}
% Human data scientists often apply a two-step decision process: they first determine whether a certain type of algorithm (e.g., feature selection) is required, and if the answer is positive they attempt to determine which specific algorithm. We opt for a similar approach by 
We group our primitives into the following families: (1) data preprocessing: data cleaning, feature encoding, etc.; (2) feature preprocessing: feature discretization, scaling and normalization; (3) feature selection: uni-variate selection, entropy-based selection etc.; (4) feature engineering: data enrichment, dimensionality reduction etc.; (5) classification and regression models: XGBoost, Random Forest, lasso regression, etc., and; (6) Combiners: the same algorithms as the previous family, but here they are used as meta-learners, combining the prediction results produced by the primitives of family 5. Their input is not limited to predictions only, and can also receive as input the output of other types of primitive families.

As we later explain in detail, at any time-step our agent can only select from a single family of primitives. 
By doing so we reduce the sizes of both the state space and the action space of the agent.

\subsubsection{Grid-world representation of the pipeline}

We design our pipeline representation based on two common practices used by data scientists and academic studies. The first is the specific order that primitive families are applied in a pipeline (e.g., pre-processing $\rightarrow$ feature selection $\rightarrow$ classification) and the second is the union of inputs from two or more sub-pipelines into a single pipeline (e.g., in \cite{he2017neural}).

As depicted in Figure \ref{fig:GridPipeline}, our state space is defined as a 6x$N$ grid, consisting of six columns -- one for each of the six primitive families -- and $N$ rows. Each cell of the grid is a placeholder for a primitive. The cells of a given column can only contain members of that column's assigned primitives family. Grid cells can be left empty, meaning that a pipeline doesn't have to contain all types of primitives.

The output of a grid cell is automatically passed to the subsequent non-empty cell in the same row. For example, in Figure \ref{fig:GridPipeline} we see that the Mutual Information component (step \#2) is connected to the XGBoost classifier (step \#3) since the Feature Engineering cell was left empty. Additionally, the output of a cell can be passed to additional cells in other rows, thus creating more complex pipelines. A cell can be connected to any other cell under two conditions: (i) that a cell's input is \textit{valid}, with predefined rules (see the following section and Alg. \ref{Alg:openlist}), and (ii) the column index of the target cell is equal or larger than that of the source cell. Any cell may receive multiple inputs, in which case all inputs are concatenated and duplicate columns are removed. 
An example of this is presented in Figure \ref{fig:GridPipeline}, where the RF classifier (step \#6) receives inputs from two cells (steps \#1 and \#5).

We further reduce the size of the state space by constraining the transition between states. Only a single grid cell -- marked by the dot cursor in Figure \ref{fig:GridPipeline} -- can be updated at any time step. Additionally, the cursor performs only a single pass on the grid in each episode, completing one row in the grid before moving to the next, making the process more manageable and efficient. 
% In addition to making the pipeline generation challenge more manageable, this approach significantly improves running time.

\begin{comment}
\subsubsection{The \yuval{Combiner} Column}
Our representation of the environment may result in multiple prediction primitives (algorithms), each producing their own classification prediction.  Moreover, it is possible for our agent to produce a linear pipeline without a prediction primitive. To produce a single unified output for the entire pipeline, we added the Ensemble column which receives the outputs of all linear pipelines and can be populated by primitives from the meta-estimator family. 

The ensemble column consists of a single cell, and it has to be populated if two or more disjoint pipelines exist. Consider the example in Figure \ref{fig:GridPipeline}: since the presented grid contains two prediction primitives (\#3 and \#6) as well as a data preprocessing primitive (\#7) without any connections, the ensemble column will be populated by an primitive that receives the output of the three primitives as input. \gilad{I know this is not completely accurate. We'll need to revise}\lior{more details are needed here? do you combine the output - concatenation?}
\yuval{Since we added the combiner as an additional family, I think we can remove this subsection - the combiner cells are just like any other cell}.
\end{comment}

\subsubsection{State Representation} Although our generated pipelines may vary in width and depth, we create a fixed-size representation that can be easily used as input for a neural net (NN). The state vector is a concatenation of the following:
\begin{itemize}
    \item  $\overline{G}_{p}$ -- represents the primitives in the grid. We use a one-hot encoding for the primitives of each cell (blank cells have their own encoding value), so the length of this vector is $N_{cells}X|\mathcal{P}_r|$, where $N_{cells}$ is the number of grid cells. This vector is sparse, and we create an embedding to compress it (see the following section and Figure \ref{fig:network}).
    \item $\overline{G}_{in}$ -- represents the incoming edges (i.e., inputs) of each cell in the grid. The length of this vector is $N_{in}XN_{cells}$, where $N_{in}$ is a configurable parameter defining the maximal number of inputs per cell. For example, in Figure \ref{fig:GridPipeline}, the cell containing the RF classifier (\#6) has two inputs (\#1 and \#5). If $N_{in}=3$, then the vector entry for this grid cell would be $\{1,5,-1\}$.
    \item $\overline{P}_{m}$ -- general meta-data describing the pipeline's topology: number of nodes and edges, graph centrality etc.
    \item $\overline{O}_{m}$ -- dataset-based features, representing the data being processed by the current grid cell. The values of this vector include the number of features, number of instances, percentage of numeric features etc. It is important to note that we generate the representation for the \textit{current form} of the dataset after being processed by the current primitives in the pipeline.
    \item $\overline{L}_{j}$ -- concatenation of the following vectors: a \textit{task} vector, with one entry for each possible task (e.g., regression, classification, etc.) that can be pursued by our model, a \textit{metric} vector, with one entry for each possible metric (e.g., accuracy, AUC, etc.) and a dataset-based features vector of the raw dataset, similarly to $\overline{O}_{m}$.
    \item $\overline{A}_{c}$ -- a vector representing the available actions. Each action is represented by a vector that details both a candidate primitive to the current grid cell, in addition to its connection(s) to the cells that provide its input. Each time step, $n$ actions are available for choice, as shown in the bottom row of figure \ref{fig:GridPipeline}. 
    All $n$ action vectors are concatenated. We elaborate on this further in the next section.
    
\end{itemize}

\begin{algorithm}[t]
\SetAlgoLined
\DontPrintSemicolon
\KwInput{$N_{in}$ - max inputs allowed, $(r, c)$ - cursor's coordinates where \(r\in 1,...,N\) and \(c \in 1,...,6\), $l$-column index of last populated cell in row $r$ which is not "blank", $G$-the grid world, $\mathcal{F}_{r,c}$ - set of family primitives associated with cell $G_{r,c}$}
  \KwOutput{\textit{open list}}

    \SetKwFunction{FMain}{OpenList}
    \SetKwProg{Fn}{Function}{:}{}
    \Fn{\FMain{$G$, $N_{in}$, $r$, $c$, $l$, $\mathcal{F}_{r,c}$}}{
    $ open\ list  \longleftarrow \emptyset$ \\
        % $ \mathcal{O}  \longleftarrow \emptyset$
        \tcp{Generate previous Outputs set}

%         \For{$ i \in [1,r) $}
%         {
%         \For{$j \in [1,c]$}{
%             \If{$ GW(i, j) \neq blank$}
%             {
%             \tcp{Add $O_{i, j}$, the output of pipeline-step in $GW(i, j)$ to the outputs set} 
%                 $ \mathcal{O}  \longleftarrow \mathcal{O} \cup O_{i, j} $
%             }
%         }
% }
$\mathcal{O} \longleftarrow \{O_{i, j}|i\in [1, r), j\in [1, c],G_{i, j}\neq"blank"\}$ \\
$ \mathcal{O}  \longleftarrow \mathcal{O} \cup O_{r, l} $ \\
    \tcp{Generate input candidates set} 
    $\mathcal{I} \longleftarrow \{s|s \subset \mathcal{O}, |s|\leq N_{in}, O_{r,l} \in s\}$\\
 \For{$ primitive \in \mathcal{F}_{r,c} $}
    { 
    \For{$ inputs \in \mathcal{I}$}
        { 
         \If{$CanAccept(primitive, inputs)$}
         {$s\longleftarrow PipeStep(primitive, inputs)$\\
         $open\ list \longleftarrow open\ list \cup s$
         }
        } 
    }    
$open\ list \longleftarrow open\ list \cup "blank"$\\
\textbf{return} $open\ list$ 
}
\textbf{End Function}
\caption{Actions Open List}
\label{Alg:openlist}
\end{algorithm}

\subsection{Hierarchical-Step Plugin}
% In the previous section we described how we restrict our state space by allowing only a single cell to be updated at any given time step. 
We first describe how we generate the set of possible actions for each time step, reducing the action space of our environment. We denote this set as the \textit{actions open list}. We then describe our novel \textit{hierarchical step plugin} approach for reducing the actions space and accelerating training.

\subsubsection{Open List of Actions} The list of possible actions for a given cell (the one with the cursor) is determined by two elements: (1) the primitive family assigned to the cell, and; (2) the set of possible candidate inputs, out of all the grid cells' outputs.

The process for generating the open actions list is presented in Algorithm \ref{Alg:openlist}. We begin by creating the set of all possible combinations inputs, denoted by $\mathcal{I}$. While a cell $C_{rc}$ in row $r$ and column $c$ always receives as input the output of the most recently populated cell in row $r$, it may also receive up to $N_{in}-1$ additional inputs from any other previously populated cell with an equal or smaller column index in previous rows. For example, step \#6 in Figure \ref{fig:GridPipeline} will always receive input from step \#5, but may also receive inputs from step \#1, \#2 and \#3. step \#4 cannot be an input for \#6 because it precedes \#5, and \#7 also cannot be used because it is populated at a later time step.

Once we crated the set of all possible inputs, we match every item in this set to all the members of the cell's assigned family primitives (lines 6-10 in Alg. \ref{Alg:openlist}). The validity of each combinations is then examined, with possible reasons for elimination including inability to process categorical features, missing entries, or negative values. All valid combinations are retained and make up the final open list.

While our representation of the possible actions is concise, the fact that it is \textit{dynamic in size} is problematic. Because each cell is likely to have a different number of actions (and also a different set of actions), all DRL algorithms that rely on a fixed action space cannot be applied on our representation. Such algorithms include the popular policy gradients, e.g., TRPO \cite{schulman2015trust}, and deep Q-networks (DQN) \cite{mnih2015human}. While solutions to this problem exist in the literature, they are not without their shortcomings. One approach is to create a fixed-size set with all possible inputs-primitives combinations. The main disadvantage of this solution is the large size of the actions vector, which will make training the agent slow and difficult. Another option is to calculate the Q-value of every state-action pair in each iteration, but this approach is both computationally expensive and only applicable to some RL methods. For these reasons, we now propose our hierarchical approach for dynamic actions modeling.

\subsubsection{Hierarchical Representation of Actions}
Our goal is to enable a RL agent to model a varying number of actions using a fixed-size representation.
We devise a hierarchical representation of the open actions list, where each level of the hierarchy is split into equal sized clusters of the actions, defined by parameter $n$. The agent iterates over the clusters of each level, selecting one action per cluster. The chosen actions are passed to the next level of the hierarchy, which is also clustered in its turn. The process is then repeated until it reaches a hierarchy level in which there are $n$ actions at most, where one single action is chosen out of the $n$ finalists.

\begin{figure}[t]
\centering
\includegraphics[width=0.82\columnwidth]{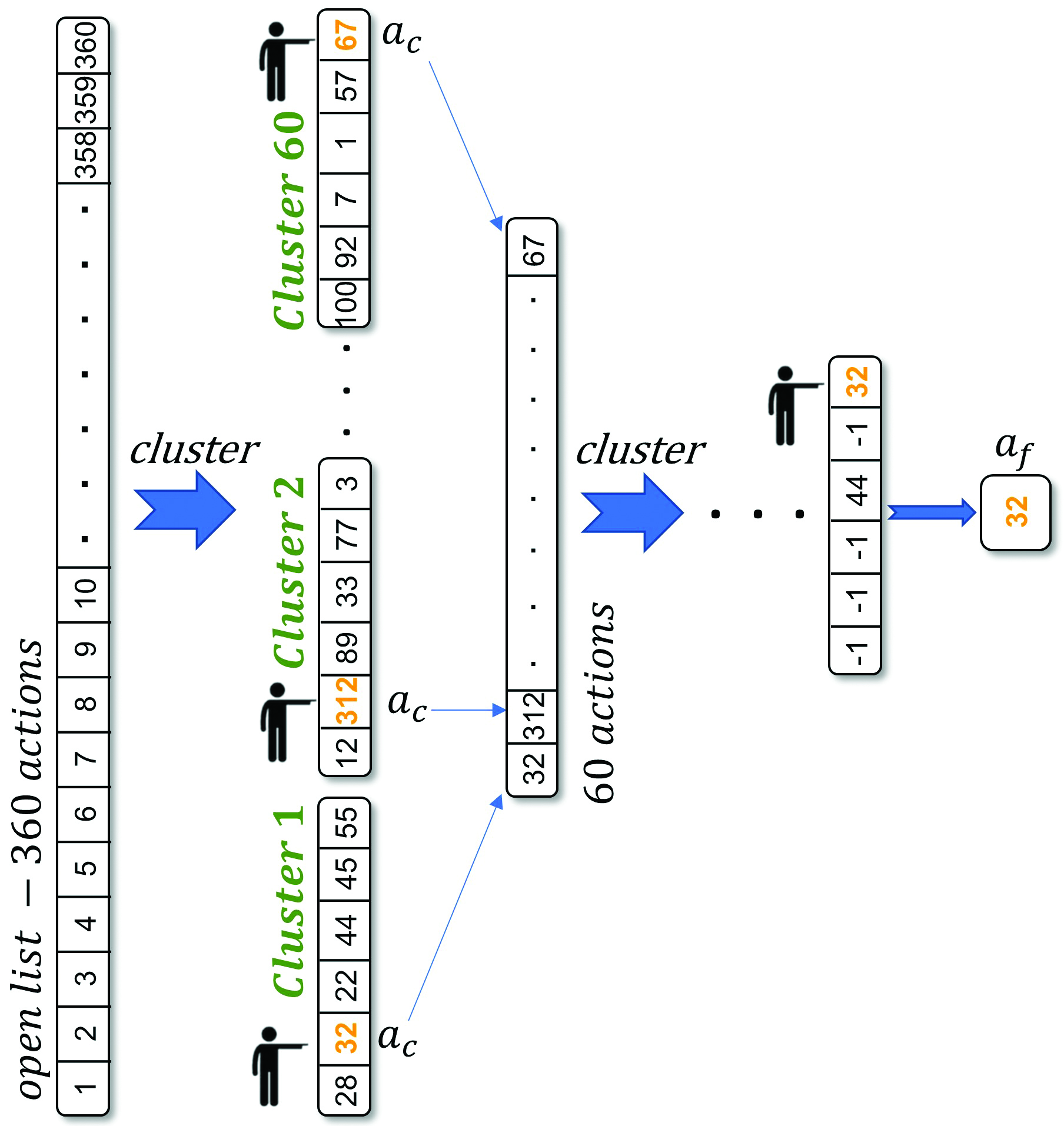} % Reduce the figure size so that it is slightly narrower than the column.
\caption[hierarchical process example]{Example of the hierarchical step process, with $n$=6 actions in each cluster and an initial open list with 360 actions.}
\label{fig:hierarchicalStep}
\end{figure}

Figure \ref{fig:hierarchicalStep} depicts an example of the process. In this example our open actions list consists of 360 possible actions. given that $n=6$, the top level of the hierarchy is split into $360/6=60$ clusters from which 60 actions are selected. 
% Note that in case the number of actions in a cluster is lower than $n$, we use padding.

\begin{comment}
We devise a hierarchical representation (a tree) of the open actions list, where each node consists of $n$ actions and has $n$ descendants. The agent iterates over the leaves of the tree, selecting one action per leaf. The chosen actions are passed to the parent nodes, which now are now populated by $n$ actions each. The process is then repeated until it reaches the root node, where one single action is chosen out of the final $n$ finalists.

An example of our approach is presented in Figure \ref{fig:hierarchicalStep}. In this example 
our tree has $7,800/6=1,300$ leaves. The agent selects one action from each leaf and passes them to the parent nodes, of which we have $1,300/6=216.7\approx217$ )(with padding). The process continues over XX layers until the final six possible actions reach the root node. 
\end{comment}

The proposed algorithm is presented in Algorithm \ref{Alg:hstep}, where $A$ and $n$ denote the actions matrix and the number of actions in each cluster, respectively. $A$ consists of vector representations of the actions in the open list, where each action vector is the concatenation of the candidate primitive vector, the inputs indices for the primitive and the dataset-based features of the input. We partition $A$ with the \textit{MakeClusters} method which returns the actions indices of each cluster. This method also pads clusters that do not have exactly $n$ actions with \textit{invalid} actions. Choosing an invalid action will prompt a negative reward (i.e., penalty). Our agent evaluates all available actions in each cluster, relying on their representations which are gathered in vector $\overline{A}_{c}$. As far as the agent concerns, it only sees a single cluster at each time-step, represented by $\overline{A}_{c}$ which is added to the state vector in line 6 of the algorithm. The selected actions are passed to $A^h$, the next level of the hierarchy. 

We create the clusters using a variation of the popular K-Means clustering algorithm \cite{hartigan1979algorithm}. The algorithm is applied at each level of the hierarchy, with the number of clusters determined by $A$ and $n$. Our approach has one significant difference from the standard algorithm, since we require not only a fixed number of clusters but also a fixed number of samples in each cluster. We achieve this by selecting for each centroid the $\frac{|A|}{n}$ nearest samples, using the cosine similarity metric. 

An important strength of the proposed approach is the fact that the hierarchical process is transparent to the agent. In addition to being compatible with a variety of popular DRL approaches, including actor-critic methods such as A3C \cite{mnih2016asynchronous} and the different variations of DQN, \MethodName places no limitations on the use of exploration-exploitation techniques such as $\epsilon$-greedy or prioritized experience replay \cite{schaul2015prioritized}, which we use in our model. Furthermore, we implement the hierarchical step as a part of our environment in compliance with OpenAI-Gym's settings, suitable for use with any DRL agent.

As we show in the evaluation section, the hierarchical approach outperforms a model with no hierarchical plugin, where the agent has to learn the environment with all possible unique actions at every time-step.

\begin{algorithm}[t]
\SetAlgoLined
\DontPrintSemicolon
\KwInput{$A$ - matrix containing dense vector of each action of the \textit{open list}; $\overline{S}$ - current state vector (without actions vector); $n$ - size of each cluster}
  \KwOutput{$\overline{S}_c, a_f$-final action $,\overline{S'}$-next state, reward, done}

    \SetKwFunction{FMain}{HierarchicalStep}
    \SetKwProg{Fn}{Function}{:}{}
    \Fn{\FMain{$A$, $\overline{S}$, $n$}}{
        $ A^h  \longleftarrow \emptyset $\\
        
        $ Clusters  \longleftarrow MakeClusters(n, A) $

        \ForEach{$ C \in Clusters $}
        {
        $ \overline{A}_c  \longleftarrow (A[C_0], A[C_1],...,A[C_n]) $ \\
        
        $ \overline{S}_c  \longleftarrow (\overline{S}, \overline{A}_c) $\tcp*{Concatenation}
        $ a_c  \longleftarrow Agent Action Selection(\overline{S}_c)$\\
        $ A^h  \longleftarrow A^h \cup A[a_c]$\\
}
        
        \eIf{$ length(Clusters)=1$}
            {
           $ a_f \longleftarrow a_c $\\
           $\overline{S'}, reward, done\longleftarrow EnvStep(a_f)$}
            {$HierarchicalStep(A^h, \overline{S}, n)$}

        \textbf{return} $\overline{S}_c, a_f, \overline{S'}, reward, done $ 
}
\textbf{End Function}
\caption{Hierarchical Step}
\label{Alg:hstep}
\end{algorithm}

\subsection{DRL Agent}

We implement our agent using the DQN algorithm, which is an off-policy algorithm. While on-policy algorithms such as policy gradients are generally more stable, they are also less sample-efficient and prone to converge to a local optimum. Moreover, while on-policy approaches generally outperform off-policy approaches in large action spaces, our hierarchical representation of actions makes this point irrelevant.

A recent improvement to the DQN algorithm is dueling-DQN (D-DQN) \cite{wang2015dueling}. D-DQN achieves faster convergence by decoupling the Q-function to the value function of the state and the advantage function of the actions, thus enabling the DQN agent to learn the value function $V(s)$, separately from the actions.
% \[Q(s,a)=V(s)+A(s,a)-\frac{1}{|\mathcal{A}|}\sum_{a=1}^{|\mathcal{A}|} A(s,a)\]

The D-DQN architecture consists of two separate sub-architectures -- one for the value function and one for the advantage of each action over the average -- each with its own output layer. Both sub-architectures are fed to a global output layer 
which computes the combined loss.

% The two output layers are then fed to a global output layer and the objective function is used to calculate the loss. The loss is then backpropagated to each of the local output layers, and then to the respective layers of each sub-architecture. This implementation can lead to large differences in the updating of the sub-architectures: for example, a ``bad'' action in a ``good'' state will result in large update to the Q-function sub-architecture and smaller updates to the value-function sub-architecture.

\begin{figure}[t]
	\centering
	\includegraphics[width=0.66\columnwidth]{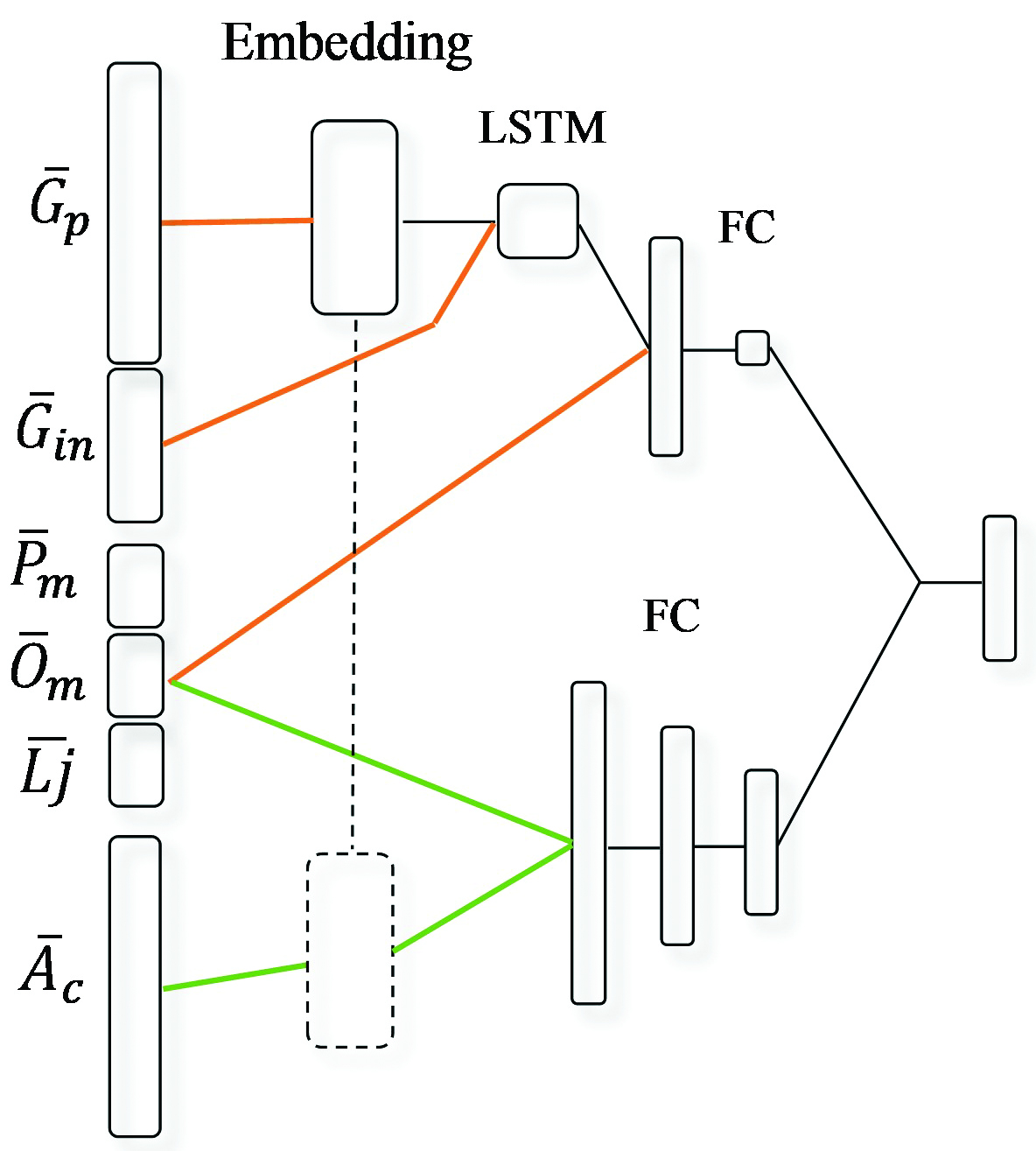}
	\caption[Network Architecture]{The D-DQN agent's NN architecture. The top stream is the \textit{value function component} and the bottom stream is the \textit{actions advantage function component}.} 
	\label{fig:network}
\end{figure}

Our implementation is a variation of the D-DQN, which makes use of the fact that our state representation consists of multiple components. Our dueling architecture is presented in Figure \ref{fig:network}. We partition the state vector as follows: the vectors that model the state of the grid form the input to the value-function sub-architecture. The vectors that model the task and the possible actions form the input to the action advantage sub-architecture. We define the architecture's objective function as follows:
\[Q(s,a)=V(s^{state})+A(s^{act},a)-\frac{1}{|\mathcal{A}|}\sum_{a=1}^{|\mathcal{A}|} A(s^{act},a)\]
where the state and action vectors are $S^{state}= \{\overline{G}_{p}, \overline{G}_{in}, \overline{P}_{m}, \overline{O}_{m},  \overline{L}_{j}\}$ and $S^{act}=\{\overline{P}_{m}, \overline{O}_{m}, \overline{L}_{j}, \overline{A}_{c}\}$.

Since $\overline{G}_p$, the vector representation of the primitives, is sparse, we add an embedding layer. Because both the action advantage sub-architecture and the hierarchical plugin use vector $\overline{A}_c$ that also contains a representation of primitives, we use the same embedding layer in all cases. Applying the same embedding for the hierarchical step means that in the early stages of the training, the actions representations are random but as time progresses the representation becomes meaningful and the clusters are more concise.

Due to the unique characteristics of our problem domain, our D-DQN implementation differs from the one proposed in \cite{wang2015dueling} in several important aspects. Most significantly, we use a long short-term memory (LSTM) architecture in the value-function sub-architecture \cite{hochreiter1997long}. We use LSTM due to the sequential manner in which we construct our ML-pipeline, where a single fixed-order sweep of the grid is performed. As a result, the action-advantage sub architecture, which consists only of fully-connected layers, is completely separate from the value-function sub-architecture. This is unlike the original D-DQN implementation, where the lower layers are shared.

Our algorithm for training the agent is identical to the one presented in \cite{mnih2015human}, except for one main difference: our use of the hierarchical-step plugin, which replaces the application of a conventional training step. However, this change, as explained in the previous section, is transparent to the architecture and does not require any modification to the D-DQN algorithm or to the exploration methods.

\section{Evaluation}
% \gilad{change the name of the section to Evaluation. the subsection should be experimental setup. The results and discussion should be another sub-section}
%In our evaluation we want to examine two things. First, the performance of our framework in the supervised classification problem in regard to its ability to produce ML pipelines compared to other methods. Here we also consider the number of pipelines evaluated during the exploration process. Secondly, we want to evaluate the effect of the hierarchical-step plugin on the convergence properties of the agent.

\subsection{Experimental Setup}
% For evaluating the main objective of our framework, pipeline generation, we conducted the following experiment. 

\subsubsection{Datasets} 
We evaluate our framework over 56 classification datasets with large variety in size, number of attributes, feature type composition and class imbalance. All datasets are available in the following online repositories: UCI,  OpenML, and Kaggle.
\subsubsection{Baselines}
We evaluate two groups of baselines:
% \begin{itemize}
    (1) \textbf{Basic popular pipelines} -- used to evaluate whether our approach is better than popular algorithms that are often used by non-experts. This group consists of three pipelines, each consisting of two pre-processing primitives -- missing values imputation  and  one-hot encoding for categorical features -- and one of the following classifiers: Random Forest \cite{liaw2002classification}, XGBoost \cite{chen2016xgboost} and Extra-Trees \cite{geurts2006extremely}, and; (2) 
    \textbf{Pipeline generation frameworks} -- we chose two popular open source pipeline generation platforms:  TPOT and Auto-Sklearn. The former achieves current state-of-the-art results, while the latter is part of one of the most popular machine learning libraries.
% \end{itemize}

\noindent For both TPOT and Auto-Sklearn, we used the default parameter settings. In the case of TPOT, this results in the generation and evaluation of 10,000 pipelines for each dataset. We run Auto-Sklearn for 30 minutes on each dataset, resulting in approximately 700 pipelines per dataset. 
% While the number of pipelines generated by Auto-Sklearn is smaller, it is important to note that the platform also performs hyperparameter tuning (which is not done by TPOT and \MethodName)\yuval{.}. 
To ensure fair comparison, we limit the list of primitives used by \MethodName to those used by TPOT.

\subsubsection{Pipeline selection} Both TPOT and Auto-Sklearn return by default a single pipeline. This pipeline is chosen based on its average performance on the folds of the training set. For \MethodName, we found that integrating our agent's Q-function into the selection process improved its performance. We define the score of a pipeline as: 
\[Score = \beta Q(S^{final}, a^{final}) + (1-\beta)KScore\]
% \gilad{Alpha is usually used for the learning rate. We need another letter}
Where \(S^{final}, a^{final}\) are the final state and action of the episode, $\beta$ is a tunable  parameter and $KScore$ is the k-fold validation performance on the training set.

We evaluate two versions of \MethodName. The first one, denoted by $v$ (for vanilla) in Table \ref{table:pipe generation results} returns the top-ranked pipeline. The second one, denoted by $e$ (for ensemble) generates $K$ pipelines and creates a weighted average of their predictions. The weight assigned to each pipeline is determined by its score.

\subsubsection{Settings} We used the following settings throughout the evaluation:
% \begin{itemize}
    We set the parameters $N$ and $n$ to 3 and 6 respectively, meaning that we use a 3X6 grid (see Figure \ref{fig:GridPipeline}) to represent our generated pipelines.
    
    The DRL agent's NN architecture is constructed as follows: the value-function
    sub-architecture consists of embedding vectors of size 15 and an LSTM of size 80, followed by three fully connected layers with lengths of 256, 128 and 32. The action-advantage sub-architecture consists of four fully connected layers with lengths of 256, 128, 64 and 32. The NN's learning rate is set to $\alpha=0.0005$.
    
    Our 56 datasets were randomly partitioned to four folds of 14 datasets each. We used K-fold cross validation: for each evaluated fold, we trained our model on the remaining three.
    Each dataset in the test fold is evaluated as follows: the dataset is split to train and test sets in a ratio of 0.8:0.2. The train set is used for the pipeline exploration by the trained agent. The returned top pipeline(s) is then evaluated on the test set with the accuracy metric.
% \end{itemize} 

\subsection{Evaluation Results and Analysis}
Table \ref{table:pipe generation results} shows the results of our evaluation. In addition to calculating the accuracy, we also present the percentage of datasets in which \MethodName's accuracy was better or equal (BOE) to that of the corresponding baseline. It is clear that both versions of \MethodName outperform all the baselines for $K\geq15$ pipelines. Moreover, the ensemble version of our approach outperforms all the baselines -- both in terms of accuracy and percentage of datasets positively affected -- by a considerable margin. 

We used paired t-test to determine whether the differences between \MethodName and the baselines are significant. Table \ref{table:p-values} shows that the difference between the ensemble version of \MethodName and all baselines are significant with $p\leq0.05$. The only exception is the ensemble version of Auto-Sklearn, for which we reach $p<0.1$. The results for the vanilla version of \MethodName are less significant, but they reach $p\leq0.15$ for all baselines.

\begin{figure}[t]
\centering
\includegraphics[width=0.9\columnwidth]{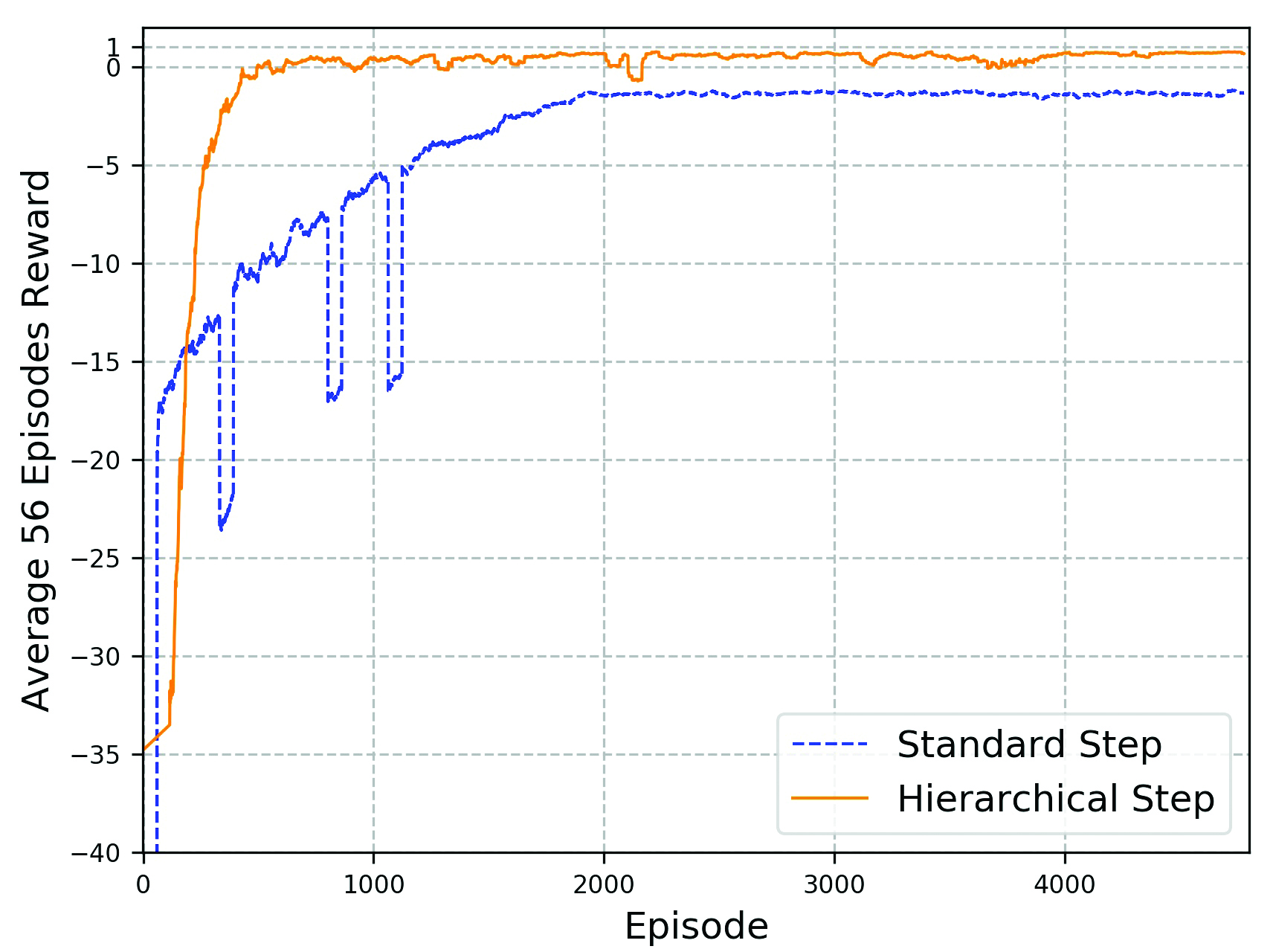} % Reduce the figure size so that it is slightly narrower than the column.
\caption[Hierarchical Step Convergence]{A comparison of our agent's convergence with and without the hierarchical step representation}
\label{fig:exp1}
\end{figure}

\subsubsection{Analyzing the contribution of the hierarchical plugin}

In order to evaluate the contribution of the hierarchical plugin, we retrain our agent using the same parameters and dataset fold partitions, but without the hierarchical representation. By removing the plugin, we increase the size of the action space available to the agent from 6 to 7,800. Figure \ref{fig:exp1} plots the average reward obtained by the agent over all 56 datasets for the first 5,000 episodes of the training. It is clear that the hierarchical plugin not only enables faster convergence, but also produces better training results.

Additional analysis of the agent's behavior during training with and without the plugin shows significant differences in action selection. While the hierarchical plugin enables the agent to explore various primitive combinations, the large action space made it difficult for the agent to effectively explore the actions space. As a result, the agent chose the ``blank'' option much more frequently. While this action is always legal (i.e., doesn't incur a penalty), it does not provide much useful information to the agent in the long term. In other words, the hierarchical plugin forces \MethodName to explore multiple actions and get useful feedback while the non-hierarchical representation delays useful exploration.

\begin{table}[t]
\small
\centering
\caption{Pipeline generation results. The result column is the average \textit{accuracy} score over 56 datasets and BOE stands for the percentage of better-or-equal results of our model (vanilla ($v$) or ensemble ($e$)) over the specified method.}
\label{table:pipe generation results}
\begin{tabular}{|p{3.5cm}|c|c|c|}
\hline
\textbf{Method} & \textbf{Result} & \textbf{BOE$^{v}$} & \textbf{BOE$^e$}  \\ \hline \hline
Random Forest   & 0.785           & 0.66& 0.75                     \\ \hline
XGBoost   & 0.791           & 0.51 & 0.607                          \\ \hline
Extra Trees   & 0.778           & 0.696 & 0.714                            \\ \hline

TPOT            & 0.793           & 0.553 & 0.589 
\\ \hline
Auto-Sklearn$^v$            & 0.784           & 0.45 & -
\\ \hline
DeepLine$^v$, $k=5$     & 0.780         & - & -                        \\ \hline
DeepLine$^v$, $k=10$    & 0.792          & -  & -                      \\ \hline
DeepLine$^v$, $k=15$   & 0.794          & - & -                       \\ \hline
DeepLine$^v$, $k=20$   & \textbf{0.799}         &  - & -                      \\ \hline
DeepLine$^v$, $k=25$  & 0.799          & - & -                       \\ \hline\hline
Auto-Sklearn$^e$ 
(top 50)
& 0.794            & - & 0.589 
\\ \hline
DeepLine$^e$, $k=5$    & 0.794          & -  & -                      \\ \hline
DeepLine$^e$, $k=10$    & 0.794          & -  & -                     \\ \hline
DeepLine$^e$, $k=15$   & 0.801           & -  & -                      \\ \hline
DeepLine$^e$, $k=20$   & 0.803         & -  & -                      \\ \hline
DeepLine$^e$, $k=25$  & \textbf{0.811}          & - & -                       \\
\hline
\end{tabular}
\end{table}

\begin{table}[t]
\small
\centering
\caption{p-values in an upper-tailed paired t-test. each cell contains the p-value obtained in the paired t-test between the accuracy results vectors of the two corresponding methods.}
\label{table:p-values}
\begin{tabular}{|c|c|c|}
\hline
 & \textbf{DeepLine$^v$} & \textbf{DeepLine$^e$}  \\ \hline \hline
Random Forest   & 0.041           & 0.002                   \\ \hline
XGBoost   & 0.151           & 0.011                           \\ \hline
Extra Trees   & 0.006           & 0.0001                             \\ \hline

TPOT            & 0.122           & 0.004  
\\ \hline
Auto-Sklearn$^v$            & 0.125            & - 
\\ \hline
Auto-Sklearn$^e$     & -         &  0.062                         \\ \hline

\end{tabular}
\end{table}

\section{Conclusions and Future Work}
We presented DeepLine, a framework for the automatic generation of ML pipelines. We use semi-constrained RL environment integrated with a novel hierarchical actions representation. our framework achieves state-of-the-art results at a much lower computational cost. 

For future work, we plan to extend our framework to include the automatic hyperparameters search and a less constrained state space.

% \section*{Acknowledgments}
% This work has been supported in part by the Defense Advanced Research Projects Agency (DARPA)
% Data-Driven Discovery of Models (D3M) Program.

\bibliography{main.bib}

\begin{thebibliography}{}

\bibitem[\protect\citeauthoryear{Anthony, Tian, and
  Barber}{2017}]{anthony2017thinking}
Anthony, T.; Tian, Z.; and Barber, D.
\newblock 2017.
\newblock Thinking fast and slow with deep learning and tree search.
\newblock In {\em Advances in Neural Information Processing Systems},
  5360--5370.

\bibitem[\protect\citeauthoryear{Bello \bgroup et al\mbox.\egroup
  }{2017}]{bello2017neural}
Bello, I.; Zoph, B.; Vasudevan, V.; and Le, Q.~V.
\newblock 2017.
\newblock Neural optimizer search with reinforcement learning.
\newblock In {\em Proceedings of the 34th International Conference on Machine
  Learning-Volume 70},  459--468.
\newblock JMLR. org.

\bibitem[\protect\citeauthoryear{Chen and Guestrin}{2016}]{chen2016xgboost}
Chen, T., and Guestrin, C.
\newblock 2016.
\newblock Xgboost: A scalable tree boosting system.
\newblock In {\em Proceedings of the 22nd acm sigkdd international conference
  on knowledge discovery and data mining},  785--794.
\newblock ACM.

\bibitem[\protect\citeauthoryear{Chen \bgroup et al\mbox.\egroup
  }{2018}]{chen2018autostacker}
Chen, B.; Wu, H.; Mo, W.; Chattopadhyay, I.; and Lipson, H.
\newblock 2018.
\newblock Autostacker: A compositional evolutionary learning system.
\newblock {\em arXiv preprint arXiv:1803.00684}.

\bibitem[\protect\citeauthoryear{Drori \bgroup et al\mbox.\egroup
  }{2018}]{drori2018alphad3m}
Drori, I.; Krishnamurthy, Y.; Rampin, R.; de~Paula~Lourenco, R.; Ono, J.~P.;
  Cho, K.; Silva, C.; and Freire, J.
\newblock 2018.
\newblock Alphad3m: Machine learning pipeline synthesis.
\newblock In {\em AutoML Workshop at ICML}.

\bibitem[\protect\citeauthoryear{Drori \bgroup et al\mbox.\egroup
  }{2019}]{drori2019automatic}
Drori, I.; Krishnamurthy, Y.; Lourenco, R.; Rampin, R.; Cho, K.; Silva, C.; and
  Freire, J.
\newblock 2019.
\newblock Automatic machine learning by pipeline synthesis using model-based
  reinforcement learning and a grammar.
\newblock {\em arXiv preprint arXiv:1905.10345}.

\bibitem[\protect\citeauthoryear{Feurer \bgroup et al\mbox.\egroup
  }{2015}]{feurer2015efficient}
Feurer, M.; Klein, A.; Eggensperger, K.; Springenberg, J.; Blum, M.; and
  Hutter, F.
\newblock 2015.
\newblock Efficient and robust automated machine learning.
\newblock In {\em Advances in Neural Information Processing Systems},
  2962--2970.

\bibitem[\protect\citeauthoryear{Geurts, Ernst, and
  Wehenkel}{2006}]{geurts2006extremely}
Geurts, P.; Ernst, D.; and Wehenkel, L.
\newblock 2006.
\newblock Extremely randomized trees.
\newblock {\em Machine learning} 63(1):3--42.

\bibitem[\protect\citeauthoryear{Hartigan and
  Wong}{1979}]{hartigan1979algorithm}
Hartigan, J.~A., and Wong, M.~A.
\newblock 1979.
\newblock Algorithm as 136: A k-means clustering algorithm.
\newblock {\em Journal of the Royal Statistical Society. Series C (Applied
  Statistics)} 28(1):100--108.

\bibitem[\protect\citeauthoryear{He \bgroup et al\mbox.\egroup
  }{2017}]{he2017neural}
He, X.; Liao, L.; Zhang, H.; Nie, L.; Hu, X.; and Chua, T.-S.
\newblock 2017.
\newblock Neural collaborative filtering.
\newblock In {\em Proceedings of the 26th international conference on world
  wide web},  173--182.
\newblock International World Wide Web Conferences Steering Committee.

\bibitem[\protect\citeauthoryear{Hochreiter and
  Schmidhuber}{1997}]{hochreiter1997long}
Hochreiter, S., and Schmidhuber, J.
\newblock 1997.
\newblock Long short-term memory.
\newblock {\em Neural computation} 9(8):1735--1780.

\bibitem[\protect\citeauthoryear{Hutter, Hoos, and
  Leyton-Brown}{2011}]{hutter2011sequential}
Hutter, F.; Hoos, H.~H.; and Leyton-Brown, K.
\newblock 2011.
\newblock Sequential model-based optimization for general algorithm
  configuration.
\newblock In {\em International Conference on Learning and Intelligent
  Optimization},  507--523.
\newblock Springer.

\bibitem[\protect\citeauthoryear{Katz, Shin, and
  Song}{2016}]{katz2016explorekit}
Katz, G.; Shin, E. C.~R.; and Song, D.
\newblock 2016.
\newblock Explorekit: Automatic feature generation and selection.
\newblock In {\em 2016 IEEE 16th International Conference on Data Mining
  (ICDM)},  979--984.
\newblock IEEE.

\bibitem[\protect\citeauthoryear{Liaw, Wiener, and
  others}{2002}]{liaw2002classification}
Liaw, A.; Wiener, M.; et~al.
\newblock 2002.
\newblock Classification and regression by randomforest.
\newblock {\em R news} 2(3):18--22.

\bibitem[\protect\citeauthoryear{Milutinovic \bgroup et al\mbox.\egroup
  }{2017}]{milutinovic2017end}
Milutinovic, M.; Baydin, A.~G.; Zinkov, R.; Harvey, W.; Song, D.; Wood, F.; and
  Shen, W.
\newblock 2017.
\newblock End-to-end training of differentiable pipelines across machine
  learning frameworks.
\newblock {\em https://openreview.net}.

\bibitem[\protect\citeauthoryear{Mnih \bgroup et al\mbox.\egroup
  }{2015}]{mnih2015human}
Mnih, V.; Kavukcuoglu, K.; Silver, D.; Rusu, A.~A.; Veness, J.; Bellemare,
  M.~G.; Graves, A.; Riedmiller, M.; Fidjeland, A.~K.; Ostrovski, G.; et~al.
\newblock 2015.
\newblock Human-level control through deep reinforcement learning.
\newblock {\em Nature} 518(7540):529.

\bibitem[\protect\citeauthoryear{Mnih \bgroup et al\mbox.\egroup
  }{2016}]{mnih2016asynchronous}
Mnih, V.; Badia, A.~P.; Mirza, M.; Graves, A.; Lillicrap, T.; Harley, T.;
  Silver, D.; and Kavukcuoglu, K.
\newblock 2016.
\newblock Asynchronous methods for deep reinforcement learning.
\newblock In {\em International conference on machine learning},  1928--1937.

\bibitem[\protect\citeauthoryear{Olson and Moore}{2016}]{olson2016tpot}
Olson, R.~S., and Moore, J.~H.
\newblock 2016.
\newblock Tpot: A tree-based pipeline optimization tool for automating machine
  learning.
\newblock In {\em Workshop on Automatic Machine Learning},  66--74.

\bibitem[\protect\citeauthoryear{Schaul \bgroup et al\mbox.\egroup
  }{2015}]{schaul2015prioritized}
Schaul, T.; Quan, J.; Antonoglou, I.; and Silver, D.
\newblock 2015.
\newblock Prioritized experience replay.
\newblock {\em arXiv preprint arXiv:1511.05952}.

\bibitem[\protect\citeauthoryear{Schulman \bgroup et al\mbox.\egroup
  }{2015}]{schulman2015trust}
Schulman, J.; Levine, S.; Abbeel, P.; Jordan, M.; and Moritz, P.
\newblock 2015.
\newblock Trust region policy optimization.
\newblock In {\em International Conference on Machine Learning},  1889--1897.

\bibitem[\protect\citeauthoryear{Silver \bgroup et al\mbox.\egroup
  }{2017}]{silver2017mastering}
Silver, D.; Hubert, T.; Schrittwieser, J.; Antonoglou, I.; Lai, M.; Guez, A.;
  Lanctot, M.; Sifre, L.; Kumaran, D.; Graepel, T.; et~al.
\newblock 2017.
\newblock Mastering chess and shogi by self-play with a general reinforcement
  learning algorithm.
\newblock {\em arXiv preprint arXiv:1712.01815}.

\bibitem[\protect\citeauthoryear{Thornton \bgroup et al\mbox.\egroup
  }{2013}]{thornton2013auto}
Thornton, C.; Hutter, F.; Hoos, H.~H.; and Leyton-Brown, K.
\newblock 2013.
\newblock Auto-weka: Combined selection and hyperparameter optimization of
  classification algorithms.
\newblock In {\em Proceedings of the 19th ACM SIGKDD international conference
  on Knowledge discovery and data mining},  847--855.
\newblock ACM.

\bibitem[\protect\citeauthoryear{Wang \bgroup et al\mbox.\egroup
  }{2015}]{wang2015dueling}
Wang, Z.; Schaul, T.; Hessel, M.; Van~Hasselt, H.; Lanctot, M.; and De~Freitas,
  N.
\newblock 2015.
\newblock Dueling network architectures for deep reinforcement learning.
\newblock {\em arXiv preprint arXiv:1511.06581}.

\end{thebibliography}
\bibliographystyle{bibs}
\end{document}